\documentclass{article}

\usepackage{arxiv}

\usepackage{hyperref}       % hyperlinks
\usepackage{microtype}      % microtypography
\usepackage{graphicx}
\usepackage{times}
\usepackage{xcolor}
\usepackage[numbers]{natbib}
\usepackage{multicol}
\usepackage{amsmath} % assumes amsmath package installed
\usepackage{amssymb}  % assumes amsmath package installed
\usepackage{float}
\usepackage{array}
\usepackage{wrapfig}
\usepackage{subcaption}
\usepackage{algorithm}
\usepackage{algpseudocode}

\title{\bf Computational Teaching for Driving via Multi-Task Imitation Learning}

\author{{Deepak Gopinath}\\
	\And
	{Xiongyi Cui} \\
        \And
        {Jonathan DeCastro} \\
        \And
        {Emily Sumner} \\
        \And
        {Jean Costa} \\
        \And
        {Hiroshi Yasuda} \\
        \And
        {Allison Morgan} \\
        \And
        {Laporsha Dees} \\
        \And
        {Sheryl Chau} \\
        \And
        {John Leonard} \\
        \And
        {Tiffany Chen} \\
        \And
        {Guy Rosman} \\
        \And
        {Avinash Balachandran}\\
        {Toyota Research Institute, Cambridge, USA}
}

\date{}

\renewcommand{\shorttitle}{Computational Teaching for Driving via Multi-Task IL}

\begin{document}
\maketitle
%%%%%%%%%%%%%%%%%%%%%%%%%%%%%%%%%%%%%%%%%%%%%%%%%%%%%%%%%%%%%%%%%%%%%%%%%%%%%%%%

\begin{abstract}
Learning motor skills for sports or performance driving is often done with professional instruction from expert human teachers, whose availability is limited. Our goal is to enable automated teaching via a learned model that interacts with the student similar to a human teacher.
However, training such automated teaching systems is limited by the availability of high-quality annotated datasets of expert teacher and student interactions that are difficult to collect at scale. 
To address this data scarcity problem, we propose an approach for training a coaching system for complex motor tasks such as high performance driving via a Multi-Task Imitation Learning (MTIL) paradigm. MTIL allows our model to learn robust representations by utilizing self-supervised training signals from more readily available \textit{non-interactive} datasets of humans performing the task of interest.

We validate our approach with (1) a semi-synthetic dataset created from real human driving trajectories, (2) a professional track driving instruction dataset, (3) a track-racing driving simulator human-subject study, and (4) a system demonstration on an instrumented car at a race track. 
Our experiments show that the right set of auxiliary machine learning tasks improves performance in predicting teaching instructions. Moreover, in the human subjects study, students exposed to the instructions from our teaching system improve their ability to stay within track limits, and show favorable perception of the model's interaction with them, in terms of usefulness and satisfaction. 
\end{abstract}


\section{Introduction}
Driving is a sensorimotor task that is done often, and requires a degree of competency that has to be taught. While daily driving is complex and safety critical, performance driving requires a higher degree of competency in handling the vehicle at high speeds and limits of stability and requires years of one-on-one instruction and practice to master. Although driving instructors can help drivers perform better and safer~\cite{isler2011effects}, their availability is limited and costly. Hence, there is a clear need for automated teaching which can help drivers improve at the population scale.

Driving instructors, e.g. in performance track driving~\cite{bentley_2019}, rely on their expertise in the driving task and their inference of student's skill levels to effectively teach students of various skill levels and learning styles. 
Instructors can gauge their students' skill levels and estimate what a student might do in a given scenario to provide contextually-relevant verbal instructions to the student. 
For example, consider how an instructor in the passenger seat might instruct a student driver on the appropriate timing for braking or the lateral positioning of the car with respect to the racing line (the optimal minimum time path around a race course). The teacher's ability to judge whether the student can maintain the racing line or oversteer in a turn influences what instructions are provided.

\begin{figure}[t]
\centering
\includegraphics[width=0.7\textwidth, height=0.45\textwidth]{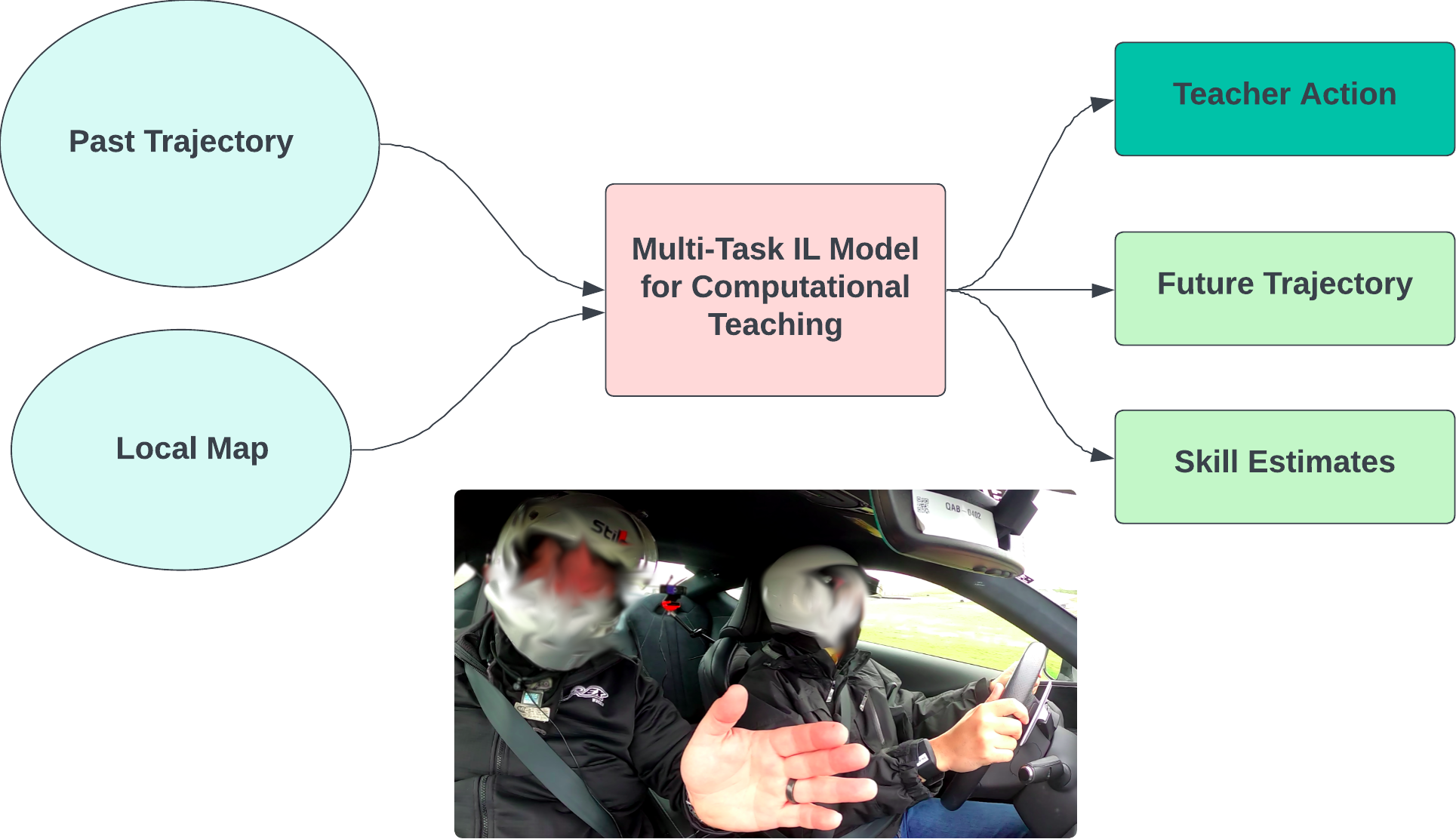}
\vspace{0.5cm}
\caption{Computational Teaching Model overview. Our model takes as input the student's past driving history as a sequence of states and controls along with local map information. The model outputs: (i) probability estimates for teacher actions (ii) future trajectories and (iii) skill estimates for the student. Our model learns to predict the correct teacher actions and can be utilized within an automated teaching system for track driving.}\label{fig:overall_schematic}
\end{figure}
An automated teaching system for driving should be able to take in relevant vehicle context (pose and dynamics, map information, etc.) and other factors (eg., driver monitoring) as inputs and output appropriate teaching actions for the student. Such a system can be trained via imitation learning, assuming access to teaching labels temporally aligned with environment states, in \textit{naturalistic} student-teacher interaction data.

Since interactive data is short in supply, the aim of this paper is to relax the need to learn from full supervision over verbal cues. We propose to leverage \textit{non-interactive} data (without student-teacher interactions) that consists of humans performing the same student tasks on their own. Such data can harness orders of magnitudes of additional supervision (e.g., from fleet data collection)~\cite{Caesar2020-uy, Ettinger_2021_ICCV,houston2021one} and enables mixing of datasets for self and semi-supervised training of imitation learned systems. This raises the important question ---

{\it{Can we attain a high-quality computational teacher by training on multiple tasks that utilize non-interactive, unlabelled data along with small amounts of interactive labeled data in a single training pipeline?}}

In this work, we seek to train and evaluate a \textit{context}- and \textit{student}-aware Intelligent Teaching System~\cite{Mousavinasab2021-lh,Nwana1990-ro} that can be deployed online for the purposes of generating driving instructions while the student is driving. 
We propose a Multi-Task Imitation Learning (MTIL) approach to address the above question. 
In our approach, we utilize an encoder-decoder structure, and our core insight is that multi-task learning with carefully selected self-supervised auxiliary tasks such as trajectory prediction~\cite{Lefevre2014-aj},
and driver performance assessment~\cite{aksjonov2018novel,chandrasiri2016driving} result in the acquisition of features and latent structures that are necessary and useful for teacher instruction prediction.

\textbf{Contributions} In this paper, we:
\begin{enumerate}
    \item Propose an approach for computational teaching based on behavior cloning in a multi-task learning setting. We probe our approach's performance and explore the relative contribution of additional self-supervisory tasks such as skill estimation and driving student behavior prediction in improving teacher action imitation.
    \item Demonstrate results on both semi-synthetic teaching data with ground-truth skill labeling with driving scenarios sampled from the Waymo dataset~\cite{Ettinger_2021_ICCV} and real-life performance driving with verbal instruction data. 
    \item Report preliminary results from a human-in-the-loop experiment in which the proposed model is used to teach how to perform track driving in a driving simulator via verbal cues provided in real time. We further demonstrate how our approach runs in real-time on an instrumented car on a race track.
\end{enumerate}

\section{Related Work}
Our approach relates to works in AI for education, motor skill learning, and multitask learning~\cite{Zhang2022-jn}, especially with respect to human-robot interactions.

Significant research efforts have been made into the development of intelligent tutoring systems in a variety of fields~\cite{Mousavinasab2021-lh,Nwana1990-ro,Swets1965-me}, from reading \cite{Atkinson1966-eb} to medicine~\cite{Rojas-Munoz2020-jd,Rojas-Munoz2021-as}, and aviation~\cite{anderson2000behavioral}, among others. Individual lines of work explored aspects such as knowledge tracing, skill estimation, and assessment~\cite{Abdelrahman2023-jg,Loh2020-xu,Piech2015-od}, and different approaches to structure the optimal policy~\cite{Keles2009-cx}, as well as jointly learning different aspects of the educational process in a multitask setting~\cite{Chaudhry2018-ry}. Other research efforts have proposed conditioning teaching actions on the current student progress level~\cite{Keles2009-cx}, modeling the teacher via approaches such as behavior cloning~\cite{moore2019behavioural} and reinforcement learning~\cite{Daubigney2013-uk,Nie2023-bt,Subramanian2021-gq}.

Several approaches looked at student modeling for the purpose of optimal teaching, either focusing on modeling the student~\cite{anderson1990cognitive,Langley-1984}, or leveraging student models within a sequential decision-making formulation of the teaching problem~\cite{Yu2023-ma}. Unlike these approaches, in our work we do not explicitly assume a student model of learning and progress, but rather implicitly capture such notions as a part of the learned latent representation within the model.
Motor learning/teaching approaches either optimally schedule exercises~\cite{Srivastava2022-mg} or provide corrective suggestions to optimize student performance~\cite{Srivastava2023-ix}

Within the human-robot interaction community, multitask learning has been leveraged in several contexts. Several works considered co-training across joint human and robot task choices ~\cite{fu2021adaptive,Huo2023-ey}, across modalities~\cite{islam2022mumu}, leveraging self-supervised data for augmenting other machine learning tasks for interaction~\cite{Liu2023-eh, Wang2023-ib}. More broadly, multitask learning has been used to build in additional context other than teaching actions~\cite{Chen2023-gl,Nguyen2023-ty,Zhang2023-kd}.
Several recent robotic efforts have focused on extracting individual skills for learning plans~\cite{Srivastava2022-mg}, adapting to a specific student style~\cite{Hou2023-rj}, and eliciting co-adaptation~\cite{Hong2023-pp}. Often, the teaching modalities are either visual or physical, as opposed to the verbal cues we are using in our work.

Finally we acknowledge a great deal of work on machine teaching \cite{zhu2018overview}, where  the learners are machine learning algorithms. The focus in such works is in attaining a certain performance of a learner in a data efficient way.

\section{Technical Approach}
In this section, we introduce the mathematical formalism, assumptions, and our specific modeling choices. 

\subsection{Problem Formulation}
\begin{figure*}[t]
\label{fig:mtl_aic_model}
\centering
\includegraphics[width=\textwidth, height=0.35\textwidth]{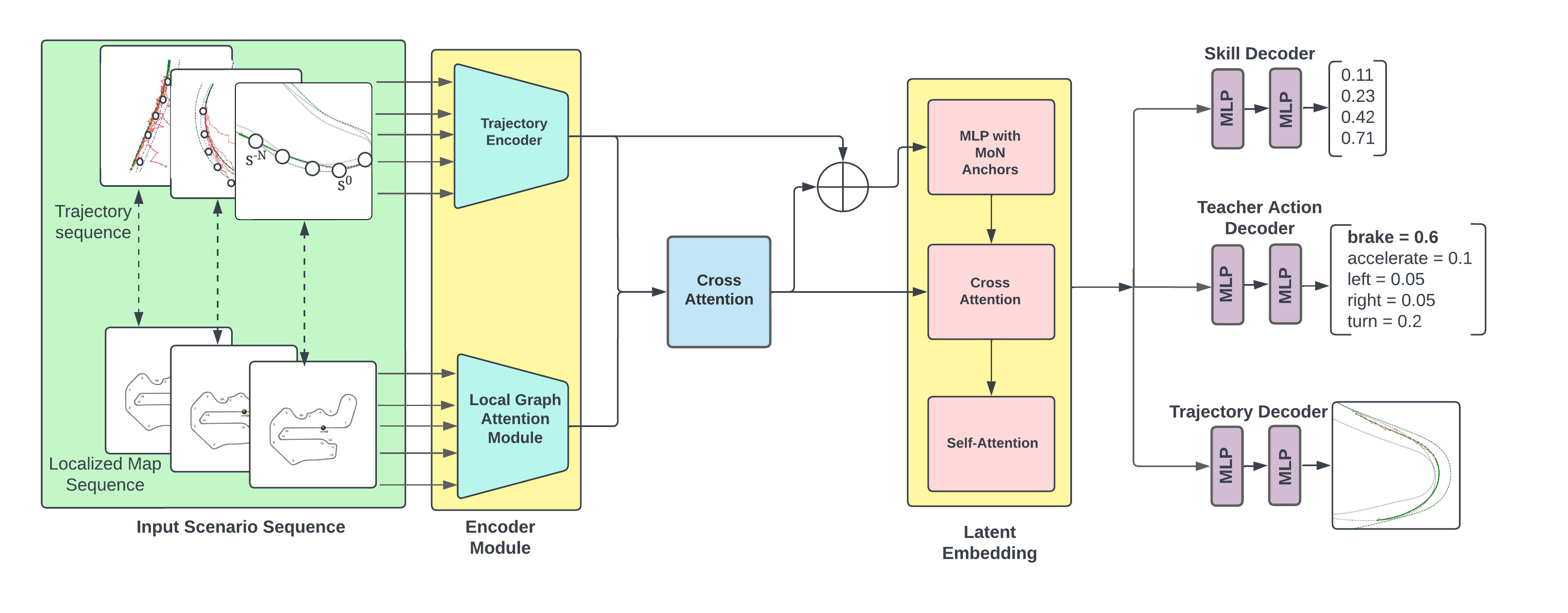}
\caption{
Model architecture for computational teaching and multi-task learning. The inputs consist of a sequence of past trajectories, $(\tau^{-N:0})^P$ encoded via a trajectory encoder, and the corresponding local map representations $m_{local} \in \mathcal{M}_{local}$ encoded via a graph attention model, along with information about the optimal racing line for experiments with racing data. The trajectory encoder is an MLP+transformer~\cite{Ngiam2022-ny} for experiments with the semi-synthetic dataset and an MLP+LSTM~\cite{alahi2016social} for the professional track driving instruction dataset. The trajectory and map embedding are fused via a cross-attention mechanism to generated an encoded state. The latent embeddings are computed by applying a series of attention based and pooling operations and then get fed to three MLP decoders for the primary (teacher action) and auxiliary (skill and trajectory) prediction tasks. \label{fig:architecture}}
\end{figure*}

Let $s^t \in \mathbb{R}^n$ denote the human-driven ego car states. Let $\tau \in \mathcal{T}$ denote a trajectory with $\mathcal{T}$ a dataset of trajectories, such that $\tau^{-N:0} = \{s^{-N}, s^{-N+1}, \dots, s^{0}\}$ is the past trajectory with $t=0$ as the current time and superscripts denote the time index. Let $\mathcal{M}_{global}$ be the global map and $\mathcal{M}_{local}$ denotes the set of \textit{localized} maps for each trajectory $\tau$ with respect to the pose of the ego car at $t=0$. We refer to the tuple $(\tau^{-N:0}, m_{local}) \in \mathcal{T} \times \mathcal{M}_{local}$ as a \textit{scenario}. 

We define the computational teaching policy as a mapping $\Pi: (\mathcal{T} \times \mathcal{M}_{local})^P \rightarrow \mathcal{A}$, where $P$ is the length of the \textit{sequence} of scenarios used to capture driving history and $\mathcal{A}$ is a set of teacher action categories that are output on the $P^{th}$ scenario in the sequence. In the context of driving, $\mathcal{A}$ could be $[\texttt{brake}, \texttt{accelerate}, \texttt{stay left}, \texttt{stay right}, \texttt{turn}]$. 

\subsection{Model Setup and Architecture}
Our computational model employs an encoder-decoder architecture that is common in motion prediction,~\cite{Varadarajan2022-yo}. The overall architecture is illustrated in Figure~\ref{fig:architecture}. We posit that the past driver behavior input to the model contains the necessary information to reliably learn a good representation that allows for skill, trajectory and teaching action prediction.

\subsubsection{Encoder}
The encoder module consists of two parallel components: a) a trajectory encoder that encodes the past trajectory $\tau^{-N:0}$ and b) a graph-based map encoder that encodes the local map $m_{local} \in \mathcal{M}_{local}$~\cite{gao2020vectornet}. We use a set of polylines to represent the left, center and right edges of each road lane to represent $m_{local}$. Each polyline consists of sequence of nodes ordered according to a prespecified lane direction and each node is a 2D vector of normalized position. The trajectory and map encodings are fused using a cross attention transformer to produce the encoded state. 
\subsubsection{Decoders}
The latent embeddings for the output decoders are generated by applying a series of MLP and transformer based operations on the encoded state. Specifically, first we add anchor layers~\cite{chai2019multipath}, followed by a series of attention based operations. Additional max pooling operations are also applied both along the sequence as well as the time dimensions. The different decoder heads use standard multi-layer perceptron (MLP) architectures. 
\subsection{Loss Function Design}
\subsubsection{Teacher action loss}
The teacher action prediction is treated as a \textbf{classification problem} over $N$ teacher action categories. 
The teacher action loss, $\mathcal{L}_{teacher}$ is a weighted Binary Cross Entropy (wBCE) with class weights corresponding to the ratio of negative to positive samples in each class computed over the entire dataset. 

\subsubsection{Trajectory Prediction Loss}
The trajectory decoder module emits $Q$ samples of $\delta x_{t}$ and $\delta y_{t}$, the change in vehicle pose for all $t \in [1, \dots, M]$. The 2D vehicle pose at $t^{th}$ time-step of the predicted trajectory is computed as a cumulative sum $\boldsymbol{x}_t$ = $\sum_{i=0}^{i=t-1}(\delta x_{i}, \delta y_{i})$.
The trajectory prediction loss is a Minimum-over-N (MoN)~\cite{Gupta2018-go,Thiede2019-sn} version of the Average Displacement Error (ADE) cost~\cite{Ettinger_2021_ICCV} and is given by  
\begin{equation}
    \mathcal{L}_{beh} = \min_{q \in [0, \dots, Q]}\frac{1}{M}\sum_{t=1}^{t=M}|\boldsymbol{x}_{q,t}-\boldsymbol{x}_{GT,t}|_2,
\end{equation}
where $Q$ is the number of predictor samples, $q$ denotes the sample number, and $\boldsymbol{x}_{\cdot,t}$ is the two dimensional pose of the vehicle at $t$ and $GT$ denotes the ground truth values.

\subsubsection{Skill estimation loss}
\textit{Skill} refers to \textit{local} track driving performance metrics such as smoothness in steering and average distance from optimal racing line, computed over a finite time horizon. It can also refer to \textit{global} driver characteristics, such as how conservative or aggressive a driver is in interactive maneuvers. The skill estimation decoder head predicts a continuous vector based skill estimate. The components of this skill vector correspond to different scalar objective metrics or driver characterization measures. The skill estimation loss, $\mathcal{L}_{skill}$ is given by
\begin{equation}
    \mathcal{L}_{skill} =(\boldsymbol{v}_{gt} - \boldsymbol{v}_{pred})^{2}
\end{equation}
where $\boldsymbol{v}_{gt}$ and $\boldsymbol{v}_{pred}$ denote the ground truth and predicted skill vector respectively. Given a dataset, $\boldsymbol{v}_{gt}$ are quantities that can be computed offline from the state information present in the trajectory data.

The overall loss function $\mathcal{L}$ is then computed as 
\begin{equation}
    \mathcal{L}_{total} = \alpha_{1}\mathcal{L}_{teacher} + 
    \alpha_{2}\mathcal{L}_{beh} + 
    \alpha_{3}\mathcal{L}_{skill}
\end{equation}
where $\alpha_1$, $\alpha_2$, and $\alpha_3$ are loss coefficients.

\section{Experiments and Results}

In this section, we describe our datasets, evaluation metrics and results from three separate experiments, two of which investigate the model performance under the multi-task setting (utilizing semi-synthetic urban and real-life professional racing datasets) and the third experiment validates the model in a human-in-the-loop driving simulator based study.

\subsection{Datasets}
\subsubsection{Semi-synthetic teaching dataset}\label{sssec:semi-synthetic}

In order to systematically explore the contribution of self-supervised auxiliary tasks in a multi-task setting in the driving domain, we curate a new semi-synthetic dataset using naturalistic driving scenarios from the Waymo Open Motion Dataset (WOMD)~\cite{Ettinger_2021_ICCV}. Each sample in our dataset is a \textit{sequence} of scenarios (each 9s long) derived from the WOMD, representing a student's driving history, with two additional novel components: a) a \textit{skill} vector associated with each sequence that captures the overall driver characteristics (how conservative vs. aggressive) and b) a single teaching instruction (treated as retrospective feedback) appropriate for the last scenario in the sequence. 

We narrow our focus on interactive maneuvers such as yields and merges. These maneuvers are categorized as \textit{aggressive}, \textit{conservative}, or \textit{neither} based on the parameter values of hand-crafted maneuver filters. We then sample $K$ \textit{scenario sequences}, each containing $P$ driving scenarios. Intuitively, a \textit{scenario sequence} represents a behavior log that captures the driver's recent history. The proportion of \textit{conservative} and \textit{aggressive} driving scenarios in the sequence is treated as proxy for overall driver skill; for example, a scenario sequence with mostly \textit{conservative} driving scenarios could be interpreted as a driver with low driving skill. A two-dimensional \textit{skill vector}, $(\alpha,\beta)$ sampled uniformly from $\mathcal{U}(0,\frac{1}{2})$, controls this proportion. We define the ground truth skill for each sequence as $\boldsymbol{v}_{gt} = (n_c, n_a)$, where $n_c$ and $n_a$ are, respectively, the number of conservative and aggressive scenarios in the sequence. 

We assign a teaching action $a_P$ from a predefined set of teaching actions $[\texttt{no\_op}, \texttt{slow\_down}, \texttt{speed\_up}]$ to the $P^{th}$ scenario in the sequence. The teacher action is a function of the label (\textit{conservative}, \textit{aggressive}) associated with the $P^{th}$ scenario as well as the skill vector ($\alpha, \beta$) associated with the sequence.
We define a hyperparameter $\gamma \in [0,1]$, that determines the relative influence of the label and the skill vector on the teacher action, with higher values of $\gamma$ indicating higher influence of the skill vector and lower influence of the label. For example, if $\gamma=0.0$ and the label for the $P^{th}$ scenario is aggressive, the teacher action is set as $\texttt{slow\_down}$; the driver is instructed to slow down and depends solely on the driving behavior in the $P^{th}$ scenario, ignoring any information from previous scenarios. If $\gamma=1.0$ and the past scenarios as a whole indicate a conservative driver, for example if $(\alpha, \beta) = (0.49, 0.05)$, then the teacher action is set to $\texttt{no\_op}$, despite an aggressive $P^{th}$ scenario. Intuitively, different values of $\gamma$ correspond to different teachers who place different levels of emphasis on \textit{immediate} driving behavior feedback (as captured by the label) vs. more global skill level (as captured by the skill vector). Lastly, we use a scenario sequence length $P=5$.

\subsubsection{Professional Track Driving Instruction Dataset}\label{sssec:thill_dataset} 
For training the teacher model for imitation of real-life track driving instruction, we collected a novel dataset of student-teacher interactions from professional track driving instruction sessions. Fourteen driving students of minimal to intermediate track driving experience received in-person track driving instruction from a professional instructor at the Thunderhill Raceway west track (Willows, CA). Students drove in a car equipped with in-cabin and vehicle data recording capabilities. The dataset consists of time synchronized vehicle data (pose, velocity, vehicle controls) and teacher instructions (recorded using microphones and transcribed using OpenAI's Whisper speech recognition and transcription tool~\cite{radford2023robust}) with poor quality transcriptions excluded manually. Each driver had approximately 30-40 minutes of instruction, and the entire dataset contains approximately seven hours of driving instruction data. 

To produce a taxonomy for the types of instructions delivered by instructors, we manually annotated a subset of the transcribed verbal utterances using the software MAXQDA. We excluded all of the student utterances and any dialogue unrelated to the driving task. This approach was intended to comprehensively document all directive commands issued by instructors. 
The taxonomy we developed included two primary categories: `vehicle', which pertained to actions influencing the vehicle's operation, such as accelerating, braking, or turning; and `driver', which related to actions involving the driver's physical movements, such as turning the head or glancing to the right. We chose to concentrate focus solely on instructions that belong to the `vehicle' category. The taxonomy consists of categories for adjusting lateral position (left, right, straight), turning (navigating a corner), accelerating, and braking. In order to categorize the entire set of transcribed sentences into proper instruction categories, we fine-tune a few-shot sentence classification model \cite{Tunstall2022-iq} using $\sim$1000 manual annotations and perform inference on the entire set of transcriptions. Manual inspection and correction of classification results was done to further improve the results. 
Since this dataset had a rich density of verbal instructions, we cast teacher action prediction as \textit{multilabel} classification. This is because, multiple teacher action categories could be present in the time horizon $[1, M]$; for example a driving coach could instruct a student to brake and right away to accelerate when entering and exiting a corner. By doing so we focus on the action category and obviate the exact timing. With this dataset the model is trained to output a vector of dimensionality $|\mathcal{A}|$, whose $i^{th}$ dimension is probability of whether teacher action category $a_{i} \in \mathcal{A}$ is present or absent in $[1, M]$. When training with this data, we use $N$=$4s$, $M$=$4s$, and $P$=$1$. This is because, in the real-life data, instructors primarily relied on recent past and the average interval of instruction was about 4s. 
The dataset has 4500 trajectories with at least one instruction category present. Lastly, when the model is deployed online, we consider the set of teacher action categories whose probabilities are above a predefined decision threshold and choose the one with the highest probability. 
\begin{table*}[t]
    \caption{\label{table_waymo}Mean and standard deviations for weighted $F_1$-score on the semi-synthetic dataset for different multi-task combinations, different values of $\gamma$, and auxiliary task dataset sizes; \textbf{A} - teacher action prediction, \textbf{AT} - teacher action + trajectory prediction, \textbf{AS} - teacher action + skill prediction, and \textbf{AST} - teacher action + skill + trajectory prediction. For each $\gamma$, the best and least performing multi-task/auxiliary task dataset size is denoted in red and blue respectively. Results in bold denote the best MTL combination for each row. We report metrics computed after early stopping averaged over eight seeds.}
    \begin{subtable}{\linewidth}
        \caption{$\gamma = 0.0$. Labelled teaching dataset size is 1.5k sequences.}
        \centering
        \begin{tabular}{|m{1.3cm}|m{2cm}|m{2cm}|m{2cm}|m{1.5cm}|}
            \hline
            Unlabelled dataset size & \multicolumn{4}{c|}{Weighted $F_1$-score (\%)}     \\ \hline
             & \multicolumn{1}{c|}{\textbf{A}}     & \multicolumn{1}{c|}{\textbf{AT}}             & \multicolumn{1}{c|}{\textbf{AS}}    & \multicolumn{1}{c|}{\textbf{AST}} \\ \hline
            0     & \multicolumn{1}{l|}{$68.9 \pm 3.7$} & \multicolumn{1}{l|}{$78.0 \pm 1.6$}          & \multicolumn{1}{l|}{\textcolor{blue}{$68.6 \pm 7.5$}} & {$\mathbf{77.5 \pm 2.7}$} \\ \hline
            10k   & \multicolumn{1}{l|}{$67.5 \pm 5.3$} & \multicolumn{1}{l|}{$80.5 \pm 1.6$}          & \multicolumn{1}{l|}{$77.5 \pm 4.1$} & {$\mathbf{82.0 \pm 1.4}$} \\ \hline
            80k   & \multicolumn{1}{l|}{$70.4 \pm 4.2$} & \multicolumn{1}{l|}{{$81.0 \pm 0.7$}} & \multicolumn{1}{l|}{$78.3 \pm 3.4$} &  \textcolor{red}{{$\mathbf{83.7 \pm 1.3}$}}  \\ \hline
        \end{tabular}
    \end{subtable}
    
    \begin{subtable}{\linewidth}
        \caption{$\gamma = 0.1$. Labelled teaching dataset size is 1.5k sequences.}
        \centering
        \begin{tabular}{|m{1.3cm}|m{2cm}|m{2cm}|m{2cm}|m{1.5cm}|}
            \hline
            Unlabelled dataset size & \multicolumn{4}{c|}{Weighted $F_1$-score (\%)}     \\ \hline
                 & \multicolumn{1}{c|}{\textbf{A}}     & \multicolumn{1}{c|}{\textbf{AT}}             & \multicolumn{1}{c|}{\textbf{AS}}    & \multicolumn{1}{c|}{\textbf{AST}} \\ \hline
            0     & \multicolumn{1}{l|}{\textcolor{blue}{$59.5 \pm 12.1$}} & \multicolumn{1}{l|}{$71.2 \pm 3.2$}          & \multicolumn{1}{l|}{$60.1 \pm 6.3$} & {$\mathbf{72.4 \pm 1.9}$} \\ \hline
            10k   & \multicolumn{1}{l|}{$63.5 \pm 3.9$} & \multicolumn{1}{l|}{$74.7 \pm 1.9$}          & \multicolumn{1}{l|}{$71.2 \pm 3.0$} & {{$\mathbf{77.0 \pm 1.2}$}}\\ \hline
            80k   & \multicolumn{1}{l|}{$63.3 \pm 3.3$} & \multicolumn{1}{l|}{{$75.0 \pm 0.8$}} & \multicolumn{1}{l|}{$73.0 \pm 2.1$} &  {\textcolor{red}{$\mathbf{77.3 \pm 0.9}$}} \\ \hline
        \end{tabular}
        
    \end{subtable}
    \begin{subtable}{\linewidth}
        \caption{$\gamma = 0.9$. Labelled teaching dataset size is 8k sequences.}
        \centering
        \begin{tabular}{|m{1.3cm}|m{2cm}|m{2cm}|m{2cm}|m{1.5cm}|}
            \hline
            Unlabelled dataset size & \multicolumn{4}{c|}{Weighted $F_1$-score (\%)}     \\ \hline
                  & \multicolumn{1}{c|}{\textbf{A}}     & \multicolumn{1}{c|}{\textbf{AT}}             & \multicolumn{1}{c|}{\textbf{AS}}    & \multicolumn{1}{c|}{\textbf{AST}}\\ \hline
            0     & \multicolumn{1}{l|}{$\mathbf{63.2 \pm 5.2}$} & \multicolumn{1}{l|}{\textcolor{blue}{$61.0 \pm 3.3$}}          & \multicolumn{1}{l|}{${62.3} \pm 4.4$} & {$61.8 \pm 4.1$} \\ \hline
            10k   & \multicolumn{1}{l|}{$63.9 \pm 3.6$} & \multicolumn{1}{l|}{$63.0 \pm 4.3$}          & \multicolumn{1}{l|}{$66.7 \pm 5.3$} & {$\mathbf{68.7 \pm 2.8}$} \\ \hline
            80k   & \multicolumn{1}{l|}{$66.8 \pm 3.7$} & \multicolumn{1}{l|}{{$64.6 \pm 5.0$}} & \multicolumn{1}{l|}{$67.8 \pm 5.2$} &  \textcolor{red}{{$\mathbf{70.9 \pm 3.8}$}} \\ \hline
        \end{tabular}
        
    \end{subtable}
    \begin{subtable}{\linewidth}
        \caption{$\gamma = 1.0$. Labelled teaching dataset size is 8k sequences.}
        \centering
        \begin{tabular}{|m{1.3cm}|m{2cm}|m{2cm}|m{2cm}|m{1.5cm}|}
            \hline
            Unlabelled dataset size & \multicolumn{4}{c|}{Weighted $F_1$-score (\%)}     \\ \hline
                  & \multicolumn{1}{c|}{\textbf{A}}     & \multicolumn{1}{c|}{\textbf{AT}}             & \multicolumn{1}{c|}{\textbf{AS}}    & \multicolumn{1}{c|}{\textbf{AST}} \\ \hline
            0     & \multicolumn{1}{l|} {\textcolor{blue}{$62.6 \pm 3.7$}} & \multicolumn{1}{l|}{$64.1 \pm 6.5$}          & \multicolumn{1}{l|}{$\mathbf{65.0 \pm 3.6}$} & {$64.2 \pm 2.4$} \\ \hline
            10k   & \multicolumn{1}{l|}{$64.2\pm 4.4$} & \multicolumn{1}{l|}{$67.3 \pm 4.5$}          & \multicolumn{1}{l|}{$70.4 \pm 2.5$} & {$\mathbf{70.7 \pm 3.2}$} \\ \hline
            80k   & \multicolumn{1}{l|}{$66.6 \pm 1.6$} & \multicolumn{1}{l|}{{$68.4 \pm 7.0$}} & \multicolumn{1}{l|}{$70.7 \pm 3.2$} &  \textcolor{red}{{$\mathbf{72.5 \pm 5.3}$}}  \\ \hline
        \end{tabular}
        
    \end{subtable}
    
\end{table*}

\subsection{Evaluation Metrics}

For modeling experiments using both datasets, we report standard multilabel (Hamming loss, weighted $F_1$-score) and multiclass (weighted $F_1$-score) classification metrics.  
For the human-in-the-loop model validation study, we consider longitudinal metrics that track a student's learning progress over multiple driving trials. We compute objective metrics that are relevant for tracking driving such as lap time and the percentage of time the driver spends outside of track bounds.

\subsection{Probing MTIL on Teaching in Daily Driving}\label{ssec:semi-synth-experiment}
We first examine the ability of our MTIL approach to handle naturalistic driving trajectories from WOMD which has rich enough diversity to allow us to evaluate the ability of our system to properly leverage skill in the provided teacher actions. To this end, we perform these experiments with the semi-synthetic dataset described in Section~\ref{sssec:semi-synthetic}. Specifically, we probe how varying levels of $\gamma$ and multi-task combinations impact teacher action prediction. The hypothesis is that higher correlation between skill and teaching action is indicative of a viable shared representation being learned from these two labels. In other words, if the teacher indeed instructs the student based on their (partial) inference of the student's skill level, then the model would perform better on teacher action prediction when it also learns to infer the underlying student skill. 
We create multiple teaching datasets with four levels of influence ($\gamma \in [0.0, 0.1, 0.9, 1.0]$) between the skill vector and the teacher action. For each $\gamma$ we also fix the size of dataset that contains the ground truth teacher labels, so as to obtain a strong training signal but not a saturated level of performance and experiment with varying sizes of the unlabelled dataset for auxiliary tasks (trajectory and skill vector prediction). This allows us to probe how relative dataset size of the primary and auxiliary tasks impacts the learning of good representations for teacher action prediction. 

Table~\ref{table_waymo} shows how weighted $F_1$-score varies for different levels of $\gamma$ for different relative sizes of dataset and multi-task combinations. For a given $\gamma$ and relative size of datasets, we observe that adding auxiliary tasks improves the classification performance when compared to single task prediction (\textbf{A} only). This indicates that in the multitask setting, the model is able to leverage shared features from additional examples so as to enhance teacher action prediction. 
In our semi-synthetic setting, adding \textit{trajectory prediction} as an auxiliary task (\textbf{AT}) has a significant impact on teacher action prediction. Although adding skill prediction alone (\textbf{AS}) improves performance, for all values of $\gamma$ and dataset sizes we find that the improvement is not as much as the \textbf{AT} condition. This is because in our dataset, driving style and behavior (conservative vs aggressive) are the primary factors that determine the teacher actions as well as the ground truth skill vector. The task of trajectory prediction enables the model to learn features that are indicative of style and behavior thereby helping in both skill prediction as well as teacher action prediction. This is seen in \textbf{AST} condition, in which both trajectory prediction and skill prediction enhance the teacher action prediction further. In Table~\ref{tab:semi_synthetic_skill_loss}, we observe that the overall skill estimation loss for \textbf{AST} condition is superior to the \textbf{AS} condition (except for $\gamma=1.0$, $10k$) indicating that trajectory prediction helps in skill estimation as well. Overall these results positively support the core premise that multi-task learning with self-supervised or self-labeled auxiliary tasks can help to enhance teacher action prediction in an imitation learning setting. 

\begin{table}
    \caption{Mean and standard deviations of skill estimation losses for $\gamma = 0.9$ and $1.0$. Note the improvement of skill given supervision from large-scale trajectory prediction.}
    \begin{subtable}{\linewidth}
        \centering
        \caption{$\gamma= 0.9$\label{tab:semi_synthetic_skill_loss_g09}}
        \begin{tabular}{|m{1.5cm}|c|c|}
        \hline
           Unlabelled dataset size  & \textbf{AS} & \textbf{AST}\\ \hline
           0  & 0.92 $\pm$ 0.25 & \textbf{0.89}$\pm$ \textbf{0.23}\\ \hline
           10k  & 0.71 $\pm$ 0.16 & \textbf{0.51} $\pm$ \textbf{0.09}\\ \hline
           80k  & 0.65 $\pm$ 0.13 & \textbf{0.55} $\pm$ \textbf{0.09}\\ \hline
        \end{tabular}
        
    \end{subtable}
    
    \begin{subtable}{\linewidth}
        \centering
        \caption{$\gamma=1.0$}
        \begin{tabular}{|m{1.5cm}|c|c|}
        \hline
           Unlabelled dataset size  & \textbf{AS} & \textbf{AST}\\ \hline
           0  & 0.99 $\pm$ 0.21 & \textbf{0.98} $\pm$ \textbf{0.21}\\ \hline
           10k  & \textbf{0.59} $\pm$ \textbf{0.04} & 0.65 $\pm$ 0.10\\ \hline
           80k  & 0.64 $\pm$ 0.14 & \textbf{0.58} $\pm$ \textbf{0.07}\\ \hline
        \end{tabular}
        
        \label{tab:semi_synthetic_skill_loss_g1}
    \end{subtable}
    
    \label{tab:semi_synthetic_skill_loss}
\end{table}

\subsection{Multi-Task Learning of Real-life Track Driving Instruction}\label{ssec:thill-experiment}

In this experiment, we utilize the professional track driving dataset described in Section~\ref{sssec:thill_dataset} to investigate our model's ability to predict teacher instructions. Additionally, we also probe how auxiliary self-supervised/labelled tasks of single agent trajectory prediction and driver performance estimation improves the baseline behavior cloning performance as a function of the size of the labelled teaching dataset. 

In Table~\ref{tbl:thill_results_bc2} we showcase the effect of multitask learning. We observed that compared to a weighted random prediction baseline (Hamming loss = $0.235$, Weighted $F_1$-score=$24.3\%$) our model is able to achieve significantly better performance (Hamming loss = $0.086$, Weighted $F_1$-score = $75.8\%$) on teacher action prediction when trained on 100\% teaching dataset. 
Compared to the urban driving scenarios, track driving is fairly constrained with well defined trajectory paths (as determined by the track geometry), and due to that, performance does not degrade drastically even when trained with relatively smaller amounts of data (Table~\ref{tbl:thill_results_bc2} rows for 50\% and 20\%). 
 
\begin{table}
\centering
\caption{Multilabel classification metrics for different task combinations and dataset sizes; Trajectory and skill prediction utilizes 100\% of the dataset ($\sim$12k snippets). Overall, multiple tasks improve teacher action prediction performance (in bold). Reported metrics are averaged over five random seeds. }
\vspace{0.5cm}
\begin{tabular}{|m{1.5cm}|lll|lll|}
\hline
Teaching Dataset Size & \multicolumn{3}{c|}{Hamming loss $\downarrow$}                                                 & \multicolumn{3}{c|}{Weighted $F_1$-score (\%) $\uparrow$}                                            \\ \hline
                      & \multicolumn{1}{l|}{A}     & \multicolumn{1}{l|}{AT}             & AST            & \multicolumn{1}{l|}{A}     & \multicolumn{1}{l|}{AT}             & AST            \\ \hline
100\%                 & \multicolumn{1}{l|}{0.086} & \multicolumn{1}{l|}{0.090}          & \textbf{0.081} & \multicolumn{1}{l|}{75.8} & \multicolumn{1}{l|}{75.6}          & \textbf{77.8} \\ \hline
50\%                  & \multicolumn{1}{l|}{0.080} & \multicolumn{1}{l|}{0.086}          & \textbf{0.075} & \multicolumn{1}{l|}{77.6} & \multicolumn{1}{l|}{76.2}          & \textbf{78.9} \\ \hline
20\%                  & \multicolumn{1}{l|}{0.087} & \multicolumn{1}{l|}{\textbf{0.081}} & 0.082          & \multicolumn{1}{l|}{74.7} & \multicolumn{1}{l|}{\textbf{76.1}} & 76.0          \\ \hline
20 examples      & \multicolumn{1}{l|}{0.222} & \multicolumn{1}{l|}{0.191} & \textbf{0.164}       & \multicolumn{1}{l|}{34.5} & \multicolumn{1}{l|}{34.5} & \textbf{38.1}          \\ \hline
\end{tabular}
\label{tbl:thill_results_bc2}
\end{table}

We notice an overall performance gain ($\sim$$1.5-2\%$) on teacher action prediction due to auxiliary task learning, except for small degradation due to the skill task in the 20\% case. In order to probe the extremely low data regime, we trained our model with 20 samples of teaching data. Unsurprisingly, the overall prediction performance drops drastically; however, the benefit of multitask learning is even higher ($\sim$4\%). 
% \vspace{-0.5cm}
\subsection{Human-in-the-Loop Validation of Model Deployment}\label{ssec:huil-experiment}

We evaluate our model's teaching efficacy in a between-subjects human-subject study, in which subjects performed a track driving task in a high-fidelity static driving simulator. 
\subsubsection{Hardware and Software} Our driving simulator utilizes CARLA~\cite{Dosovitskiy17} and ROS as the main software components. We re-created the Thunderhill Raceway west track, where the track driving dataset was collected (see Sec. \ref{sssec:thill_dataset}), adding features such as tire squeals for realism. 
\subsubsection{Study Protocol} We conducted a between-subjects study in which one group (\textbf{coaching}, $n=7$) was exposed to the teaching instructions generated by the learned model and the other group (\textbf{no-coaching}, $n=8$) performed the prescribed task on their own. Participants were recruited via user study recruiting agency Fieldwork and compensated \$150 for their two-hour participation. Upon arrival, participants signed a consent form. This research was reviewed and approved by the WCG IRB number 20232162, approval number 45594784. Participants were then randomly assigned to a group. 

Each subject initially underwent a practice phase (two laps) during which they familiarized themselves with the simulator setup and the track layout. The subjects were instructed to: a) remain on track, b) stay under 90mph, c) interpret teaching instructions as they deem fit, and d) make good lap time. After the practice phase, subjects did eight laps of training. Subjects in the \textbf{coaching} condition received teaching instructions from the model, whereas the subjects in the \textbf{no-coaching} condition did not. The subjects then drove for two laps without any teaching instructions as a retention test. 

\begin{figure}
\hfill
\centering
\begin{minipage}{0.49\linewidth}
\includegraphics[width=1.0\linewidth]{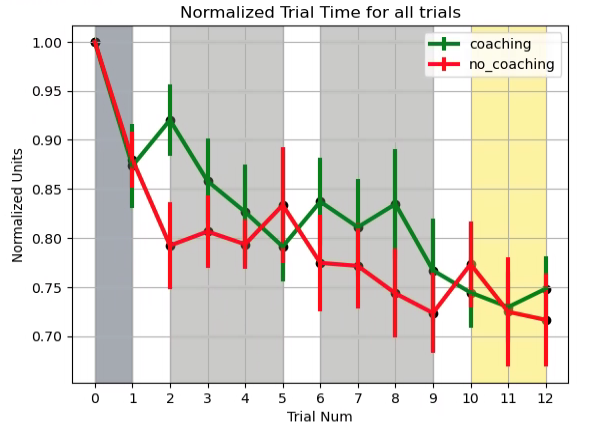}
\centering
(a)
\end{minipage} \hfill 
\begin{minipage}{0.49\linewidth}
\centering
\includegraphics[width=0.97\linewidth]{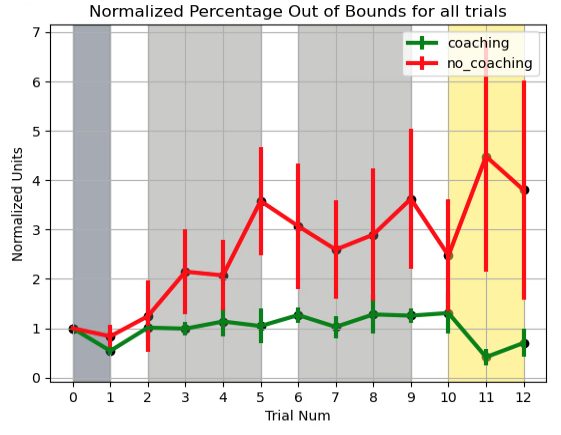}
\centering
(b)
\end{minipage}
\hfill

\caption{Evolution of objective metrics during the course of the study. (a) Normalized Lap time. (b): Normalized Percentage out of bounds. The dark gray colored band is the \textit{familiarization}, the light gray correspond to the \textit{training}, and the light yellow band to the \textit{retention} phases.}
\label{fig:huil_plots}

\end{figure}
\subsubsection{Results} In Figure~\ref{fig:huil_plots}(a) both groups exhibit an improvement in lap time during the course of the experiment. We see that the \textbf{coaching} group exhibits a sharper decrease in the lap times in the early phases of learning (trials 2, 3, 4), compared to the \textbf{no-coaching} group. Since our system primarily coaches via an action-based model, our observations are consistent with findings from the motor learning domain that action based feedback is most useful in early learning phases~\cite{fitts1967human}. In the retention phase, the \textbf{coaching} group is marginally better than the \textbf{no-coaching} group. From Figure~\ref{fig:huil_plots}(b) we find that participants in the \textbf{coaching} group are better at keeping the vehicle within track bounds. This is likely because, teaching helps the participants have better lateral vehicle positioning and brake and throttle application at appropriate times.

\subsection{Deployment Test on a Real Vehicle}\label{ssec:huil-experiment-real}
We have tested our model on an instrumented Lexus LC500 \cite{DBLP:journals/fr/DallasTGB23}, via a ROS2 subscriber module wrapping our PyTorch implementation. The model provided real-time instructions via pre-recorded verbal cues matching the emitted teacher actions, demonstrating its real-time teaching capability on real hardware platforms (see the supplementary video).

\section{Conclusions}
We demonstrated in this paper how approaches from trajectory prediction can be adopted for learning a model for driver coaching by mapping trajectory history and context into instruction verbal cues.
Within this domain, we showed how self-labeled auxiliary tasks such as trajectory and skill prediction enable successful multitask training even when labelled teaching dataset is relatively small.
We present results both on a novel semi-synthetic dataset generated based on naturalistic driving data, and based on real-life track driving instruction dataset. In a pilot study with a racing simulator, our computational teaching system helped students reduce their time and track bounds infractions, demonstrating the effectiveness of the approach and leading to new avenues of exploration for our framework.

\bibliographystyle{unsrtnat}

\begin{thebibliography}{53}
\providecommand{\natexlab}[1]{#1}
\providecommand{\url}[1]{\texttt{#1}}
\expandafter\ifx\csname urlstyle\endcsname\relax
  \providecommand{\doi}[1]{doi: #1}\else
  \providecommand{\doi}{doi: \begingroup \urlstyle{rm}\Url}\fi

\bibitem[Isler et~al.(2011)Isler, Starkey, and Sheppard]{isler2011effects}
Robert~B Isler, Nicola~J Starkey, and Peter Sheppard.
\newblock Effects of higher-order driving skill training on young, inexperienced drivers’ on-road driving performance.
\newblock \emph{Accident Analysis \& Prevention}, 43\penalty0 (5):\penalty0 1818--1827, 2011.

\bibitem[Bentley(2019)]{bentley_2019}
Ross Bentley.
\newblock \emph{The HPDE Instructor Manifesto}.
\newblock www.speedsecrets.com, 2019.

\bibitem[Caesar et~al.(2020)Caesar, Bankiti, Lang, Vora, Liong, Xu, Krishnan, Pan, Baldan, and Beijbom]{Caesar2020-uy}
Holger Caesar, Varun Bankiti, Alex~H Lang, Sourabh Vora, Venice~Erin Liong, Qiang Xu, Anush Krishnan, Yu~Pan, Giancarlo Baldan, and Oscar Beijbom.
\newblock {nuScenes}: A multimodal dataset for autonomous driving.
\newblock In \emph{CVPR}. IEEE, June 2020.

\bibitem[Ettinger et~al.(2021)Ettinger, Cheng, Caine, Liu, Zhao, Pradhan, Chai, Sapp, Qi, Zhou, Yang, Chouard, Sun, Ngiam, Vasudevan, McCauley, Shlens, and Anguelov]{Ettinger_2021_ICCV}
Scott Ettinger, Shuyang Cheng, Benjamin Caine, Chenxi Liu, Hang Zhao, Sabeek Pradhan, Yuning Chai, Ben Sapp, Charles~R. Qi, Yin Zhou, Zoey Yang, Aur\'elien Chouard, Pei Sun, Jiquan Ngiam, Vijay Vasudevan, Alexander McCauley, Jonathon Shlens, and Dragomir Anguelov.
\newblock Large scale interactive motion forecasting for autonomous driving: The {Waymo} open motion dataset.
\newblock In \emph{ICCV}, pages 9710--9719, October 2021.

\bibitem[Houston et~al.(2021)Houston, Zuidhof, Bergamini, Ye, Chen, Jain, Omari, Iglovikov, and Ondruska]{houston2021one}
John Houston, Guido Zuidhof, Luca Bergamini, Yawei Ye, Long Chen, Ashesh Jain, Sammy Omari, Vladimir Iglovikov, and Peter Ondruska.
\newblock One thousand and one hours: Self-driving motion prediction dataset.
\newblock In \emph{CoRL}, pages 409--418. PMLR, 2021.

\bibitem[Mousavinasab et~al.(2021)Mousavinasab, Zarifsanaiey, R.~Niakan~Kalhori, Rakhshan, Keikha, and Ghazi~Saeedi]{Mousavinasab2021-lh}
Elham Mousavinasab, Nahid Zarifsanaiey, Sharareh R.~Niakan~Kalhori, Mahnaz Rakhshan, Leila Keikha, and Marjan Ghazi~Saeedi.
\newblock Intelligent tutoring systems: a systematic review of characteristics, applications, and evaluation methods.
\newblock \emph{Interactive Learning Environments}, 29\penalty0 (1):\penalty0 142--163, January 2021.

\bibitem[Nwana(1990)]{Nwana1990-ro}
Hyacinths Nwana.
\newblock Intelligent tutoring systems: an overview.
\newblock \emph{Artif. Intell. Rev.}, 4\penalty0 (4):\penalty0 251--277, 1990.

\bibitem[Lef{\`e}vre et~al.(2014)Lef{\`e}vre, Vasquez, and Laugier]{Lefevre2014-aj}
St{\'e}phanie Lef{\`e}vre, Dizan Vasquez, and Christian Laugier.
\newblock A survey on motion prediction and risk assessment for intelligent vehicles.
\newblock \emph{ROBOMECH Journal}, 1\penalty0 (1):\penalty0 1--14, July 2014.

\bibitem[Aksjonov et~al.(2018)Aksjonov, Nedoma, Vodovozov, Petlenkov, and Herrmann]{aksjonov2018novel}
Andrei Aksjonov, Pavel Nedoma, Valery Vodovozov, Eduard Petlenkov, and Martin Herrmann.
\newblock A novel driver performance model based on machine learning.
\newblock \emph{IFAC-PapersOnLine}, 51\penalty0 (9):\penalty0 267--272, 2018.

\bibitem[Chandrasiri et~al.(2016)Chandrasiri, Nawa, and Ishii]{chandrasiri2016driving}
Naiwala~P Chandrasiri, Kazunari Nawa, and Akira Ishii.
\newblock Driving skill classification in curve driving scenes using machine learning.
\newblock \emph{Journal of Modern Transportation}, 24:\penalty0 196--206, 2016.

\bibitem[Zhang and Yang(2022)]{Zhang2022-jn}
Yu~Zhang and Qiang Yang.
\newblock A survey on {Multi-Task} learning.
\newblock \emph{IEEE Trans. Knowl. Data Eng.}, 34\penalty0 (12):\penalty0 5586--5609, December 2022.

\bibitem[Swets and Feurzeig(1965)]{Swets1965-me}
J~A Swets and W~Feurzeig.
\newblock Computer-aided instruction.
\newblock \emph{Science}, 150\penalty0 (3696):\penalty0 572--576, October 1965.

\bibitem[Atkinson and Hansen(1966)]{Atkinson1966-eb}
Richard~C Atkinson and Duncan~N Hansen.
\newblock {Computer-Assisted} instruction in initial reading: The {S}tanford {P}roject.
\newblock \emph{Read. Res. Q.}, 2\penalty0 (1):\penalty0 5--25, 1966.

\bibitem[Rojas-Mu{\~n}oz et~al.(2020)Rojas-Mu{\~n}oz, Couperus, and Wachs]{Rojas-Munoz2020-jd}
Edgar Rojas-Mu{\~n}oz, Kyle Couperus, and Juan Wachs.
\newblock {DAISI}: Database for {AI} surgical instruction.
\newblock \emph{arXiv}, March 2020.

\bibitem[Rojas-Muñoz et~al.(2021)Rojas-Muñoz, Couperus, and Wachs]{Rojas-Munoz2021-as}
Edgar Rojas-Muñoz, Kyle Couperus, and Juan~P Wachs.
\newblock The {AI}-medic: an artificial intelligent mentor for trauma surgery.
\newblock \emph{Comput. Methods Biomech. Biomed. Eng. Imaging Vis.}, 9\penalty0 (3):\penalty0 313--321, May 2021.

\bibitem[Anderson et~al.(2000)Anderson, Draper, and Peterson]{anderson2000behavioral}
Charles~W Anderson, Bruce~A Draper, and David~A Peterson.
\newblock Behavioral cloning of student pilots with modular neural networks.
\newblock In \emph{ICML}, pages 25--32, 2000.

\bibitem[Abdelrahman et~al.(2023)Abdelrahman, Wang, and Nunes]{Abdelrahman2023-jg}
Ghodai Abdelrahman, Qing Wang, and Bernardo Nunes.
\newblock Knowledge tracing: A survey.
\newblock \emph{ACM Comput. Surv.}, 55\penalty0 (11):\penalty0 1--37, February 2023.

\bibitem[Loh et~al.(2020)Loh, Chae, and Hwang]{Loh2020-xu}
Hyunbin Loh, Piljae Chae, and Chanyou Hwang.
\newblock Data efficient educational assessment via multi-dimensional pairwise comparisons.
\newblock In \emph{Educational Data Mining}, 2020.

\bibitem[Piech et~al.(2015)Piech, Bassen, Huang, Ganguli, Sahami, Guibas, and Sohl-Dickstein]{Piech2015-od}
Chris Piech, Jonathan Bassen, Jonathan Huang, Surya Ganguli, Mehran Sahami, Leonidas~J Guibas, and Jascha Sohl-Dickstein.
\newblock Deep knowledge tracing.
\newblock \emph{Adv. Neural Inf. Process. Syst.}, 28, 2015.

\bibitem[Kele{\c s} et~al.(2009)Kele{\c s}, Ocak, Kele{\c s}, and G{\"u}lc{\"u}]{Keles2009-cx}
Ayt{\"u}rk Kele{\c s}, Rahim Ocak, Ali Kele{\c s}, and Aslan G{\"u}lc{\"u}.
\newblock {ZOSMAT}: Web-based intelligent tutoring system for teaching--learning process.
\newblock \emph{Expert Syst. Appl.}, 36\penalty0 (2, Part 1):\penalty0 1229--1239, March 2009.

\bibitem[Chaudhry et~al.(2018)Chaudhry, Singh, Dogga, and Saini]{Chaudhry2018-ry}
Ritwick Chaudhry, Harvineet Singh, Pradeep Dogga, and Shiv~Kumar Saini.
\newblock Modeling {Hint-Taking} behavior and knowledge state of students with {Multi-Task} learning.
\newblock \emph{International Educational Data Mining Society}, July 2018.

\bibitem[Moore et~al.(2019)Moore, Caines, Rice, and Buttery]{moore2019behavioural}
Russell Moore, Andrew Caines, Andrew Rice, and Paula Buttery.
\newblock Behavioural cloning of teachers for automatic homework selection.
\newblock In \emph{International Conference on Artificial Intelligence in Education}, pages 333--344. Springer, 2019.

\bibitem[Daubigney et~al.(2013)Daubigney, Geist, and Pietquin]{Daubigney2013-uk}
Lucie Daubigney, Matthieu Geist, and Olivier Pietquin.
\newblock Model-free {POMDP} optimisation of tutoring systems with echo-state networks.
\newblock In Maxine Eskenazi, Michael Strube, Barbara Di~Eugenio, and Jason~D Williams, editors, \emph{SIGDIAL}, pages 102--106, Metz, France, August 2013. Association for Computational Linguistics.

\bibitem[Nie et~al.(2023)Nie, Reuel, and Brunskill]{Nie2023-bt}
Allen Nie, Ann-Katrin Reuel, and Emma Brunskill.
\newblock Understanding the impact of reinforcement learning personalization on subgroups of students in math tutoring.
\newblock In \emph{Conference on Artificial Intelligence in Education}, 2023.

\bibitem[Subramanian and Mostow(2021)]{Subramanian2021-gq}
Jithendaraa Subramanian and David Mostow.
\newblock Using deep reinforcement learning to train and evaluate instructional sequencing policies for an intelligent tutoring system.
\newblock In \emph{Workshop on {RL4ED}, {EDM}}, 2021.

\bibitem[Anderson et~al.(1990)Anderson, Boyle, Corbett, and Lewis]{anderson1990cognitive}
John~R Anderson, C~Franklin Boyle, Albert~T Corbett, and Matthew~W Lewis.
\newblock Cognitive modeling and intelligent tutoring.
\newblock \emph{Artificial Intelligence}, 42\penalty0 (1):\penalty0 7--49, 1990.

\bibitem[Langley et~al.(1984)Langley, Ohlsson, and Sage]{Langley-1984}
Pat Langley, Stellan Ohlsson, and Stephanie Sage.
\newblock A machine learning approach to student modeling.
\newblock Technical report, CMU, 1984.

\bibitem[Yu et~al.(2023)Yu, Xu, Li, and Hsu]{Yu2023-ma}
Cunjun Yu, Yiqing Xu, Linfeng Li, and David Hsu.
\newblock {COACH}: Cooperative robot teaching.
\newblock In Karen Liu, Dana Kulic, and Jeff Ichnowski, editors, \emph{CoRL}, volume 205, pages 1092--1103. PMLR, 2023.

\bibitem[Srivastava et~al.(2022)Srivastava, Biyik, Mirchandani, Goodman, and Sadigh]{Srivastava2022-mg}
Megha Srivastava, Erdem Biyik, Suvir Mirchandani, Noah Goodman, and Dorsa Sadigh.
\newblock Assistive teaching of motor control tasks to humans.
\newblock In \emph{NeurIPS}, November 2022.

\bibitem[Srivastava et~al.(2023)Srivastava, Goodman, and Sadigh]{Srivastava2023-ix}
Megha Srivastava, Noah Goodman, and Dorsa Sadigh.
\newblock Generating language corrections for teaching physical control tasks.
\newblock In \emph{ICML}, volume 202, pages 32561--32574, July 2023.

\bibitem[Fu et~al.(2021)Fu, Du, Teng, Fu, and Wu]{fu2021adaptive}
Jian Fu, Jinyu Du, Xiang Teng, Yuxiang Fu, and Lu~Wu.
\newblock Adaptive multi-task human-robot interaction based on human behavioral intention.
\newblock \emph{IEEE Access}, 9:\penalty0 133762--133773, 2021.

\bibitem[Huo et~al.(2023)Huo, Ding, Xu, Tian, Zhu, Mu, Sun, Tomizuka, and Zhan]{Huo2023-ey}
Mingxiao Huo, Mingyu Ding, Chenfeng Xu, Thomas Tian, Xinghao Zhu, Yao Mu, Lingfeng Sun, Masayoshi Tomizuka, and Wei Zhan.
\newblock Human-oriented representation learning for robotic manipulation.
\newblock \emph{arXiv}, October 2023.

\bibitem[Islam and Iqbal(2022)]{islam2022mumu}
Md~Mofijul Islam and Tariq Iqbal.
\newblock {MuMu}: cooperative multitask learning-based guided multimodal fusion.
\newblock In \emph{AAAI}, volume~36, pages 1043--1051, 2022.

\bibitem[Liu et~al.(2023)Liu, Chen, and Abbeel]{Liu2023-eh}
Yuxuan Liu, Xi~Chen, and Pieter Abbeel.
\newblock {Self-Supervised} instance segmentation by grasping.
\newblock \emph{arXiv}, May 2023.

\bibitem[Wang et~al.(2023)Wang, Mariani, Menciassi, De~Momi, and Fey]{Wang2023-ib}
Ziheng Wang, Andrea Mariani, Arianna Menciassi, Elena De~Momi, and Ann~Majewicz Fey.
\newblock {Uncertainty-Aware} {Self-Supervised} learning for {Cross-Domain} technical skill assessment in {Robot-Assisted} surgery.
\newblock \emph{IEEE Transactions on Medical Robotics and Bionics}, 5\penalty0 (2):\penalty0 301--311, May 2023.

\bibitem[Chen et~al.(2023)Chen, Tamboli, Lan, and Aggarwal]{Chen2023-gl}
Jiayu Chen, Dipesh Tamboli, Tian Lan, and Vaneet Aggarwal.
\newblock Multi-task hierarchical adversarial inverse reinforcement learning.
\newblock In \emph{ICML}, pages 4895--4920. PMLR, 2023.

\bibitem[Nguyen et~al.(2023)Nguyen, Tran, Bach, Tan, and Eiji]{Nguyen2023-ty}
Tho~Duc Nguyen, Chanh~Minh Tran, Nguyen~Gia Bach, Phan~Xuan Tan, and Kamioka Eiji.
\newblock Disentangled representation learning for generative adversarial multi-task imitation learning.
\newblock In \emph{CCRIS}, CCRIS '23, pages 76--80, New York, NY, USA, October 2023. Association for Computing Machinery.

\bibitem[Zhang et~al.(2023)Zhang, Kang, Lee, Tomlin, Levine, Tu, and Matni]{Zhang2023-kd}
Thomas~T Zhang, Katie Kang, Bruce~D Lee, Claire Tomlin, Sergey Levine, Stephen Tu, and Nikolai Matni.
\newblock {Multi-Task} imitation learning for linear dynamical systems.
\newblock In Nikolai Matni, Manfred Morari, and George~J Pappas, editors, \emph{L4DC}, volume 211, pages 586--599. PMLR, 2023.

\bibitem[Hou et~al.(2023)Hou, Yu, Hsu, and Yu]{Hou2023-rj}
Zhimin Hou, Cunjun Yu, David Hsu, and Haoyong Yu.
\newblock {TeachingBot}: Robot teacher for human handwriting.
\newblock \emph{arXiv}, September 2023.

\bibitem[Hong et~al.(2023)Hong, Levine, and Dragan]{Hong2023-pp}
Joey Hong, Sergey Levine, and Anca Dragan.
\newblock Learning to influence human behavior with offline reinforcement learning.
\newblock \emph{arXiv}, March 2023.

\bibitem[Zhu et~al.(2018)Zhu, Singla, Zilles, and Rafferty]{zhu2018overview}
Xiaojin Zhu, Adish Singla, Sandra Zilles, and Anna~N Rafferty.
\newblock An overview of machine teaching.
\newblock \emph{arXiv preprint arXiv:1801.05927}, 2018.

\bibitem[Ngiam et~al.(2022)Ngiam, Caine, Vasudevan, Zhang, Chiang, Ling, Roelofs, Bewley, Liu, Venugopal, Weiss, Sapp, Chen, and Shlens]{Ngiam2022-ny}
Jiquan Ngiam, Benjamin Caine, Vijay Vasudevan, Zhengdong Zhang, Hao-Tien~Lewis Chiang, Jeffrey Ling, Rebecca Roelofs, Alex Bewley, Chenxi Liu, Ashish Venugopal, David Weiss, Ben Sapp, Zhifeng Chen, and Jonathon Shlens.
\newblock Scene transformer: A unified architecture for predicting multiple agent trajectories.
\newblock In \emph{ICLR}, 2022.

\bibitem[Alahi et~al.(2016)Alahi, Goel, Ramanathan, Robicquet, Fei-Fei, and Savarese]{alahi2016social}
Alexandre Alahi, Kratarth Goel, Vignesh Ramanathan, Alexandre Robicquet, Li~Fei-Fei, and Silvio Savarese.
\newblock Social {LSTM}: Human trajectory prediction in crowded spaces.
\newblock In \emph{CVPR}, pages 961--971, 2016.

\bibitem[Varadarajan et~al.(2022)Varadarajan, Hefny, Srivastava, Refaat, Nayakanti, Cornman, Chen, Douillard, Lam, Anguelov, and Sapp]{Varadarajan2022-yo}
Balakrishnan Varadarajan, Ahmed Hefny, Avikalp Srivastava, Khaled~S Refaat, Nigamaa Nayakanti, Andre Cornman, Kan Chen, Bertrand Douillard, Chi~Pang Lam, Dragomir Anguelov, and Benjamin Sapp.
\newblock {MultiPath++}: Efficient information fusion and trajectory aggregation for behavior prediction.
\newblock In \emph{ICRA}, pages 7814--7821. IEEE, May 2022.

\bibitem[Gao et~al.(2020)Gao, Sun, Zhao, Shen, Anguelov, Li, and Schmid]{gao2020vectornet}
Jiyang Gao, Chen Sun, Hang Zhao, Yi~Shen, Dragomir Anguelov, Congcong Li, and Cordelia Schmid.
\newblock Vectornet: Encoding {HD} maps and agent dynamics from vectorized representation.
\newblock In \emph{CVPR}, pages 11525--11533, 2020.

\bibitem[Chai et~al.(2019)Chai, Sapp, Bansal, and Anguelov]{chai2019multipath}
Yuning Chai, Benjamin Sapp, Mayank Bansal, and Dragomir Anguelov.
\newblock Multipath: Multiple probabilistic anchor trajectory hypotheses for behavior prediction.
\newblock \emph{arXiv preprint arXiv:1910.05449}, 2019.

\bibitem[Gupta et~al.(2018)Gupta, Johnson, Fei-Fei, Savarese, and Alahi]{Gupta2018-go}
Agrim Gupta, Justin Johnson, Li~Fei-Fei, Silvio Savarese, and Alexandre Alahi.
\newblock Social {GAN}: Socially acceptable trajectories with generative adversarial networks.
\newblock In \emph{CVPR}, March 2018.

\bibitem[Thiede and Brahma(2019)]{Thiede2019-sn}
L~Thiede and P~Brahma.
\newblock Analyzing the variety loss in the context of probabilistic trajectory prediction.
\newblock \emph{ICCV}, pages 9953--9962, July 2019.

\bibitem[Radford et~al.(2023)Radford, Kim, Xu, Brockman, McLeavey, and Sutskever]{radford2023robust}
Alec Radford, Jong~Wook Kim, Tao Xu, Greg Brockman, Christine McLeavey, and Ilya Sutskever.
\newblock Robust speech recognition via large-scale weak supervision.
\newblock In \emph{ICML}, pages 28492--28518. PMLR, 2023.

\bibitem[Tunstall et~al.(2022)Tunstall, Reimers, Jo, Bates, Korat, Wasserblat, and Pereg]{Tunstall2022-iq}
Lewis Tunstall, Nils Reimers, Unso Eun~Seo Jo, Luke Bates, Daniel Korat, Moshe Wasserblat, and Oren Pereg.
\newblock setfit: Efficient few-shot learning with sentence transformers.
\newblock In \emph{{NeurIPS} {ENLSP} workshop}, 2022.

\bibitem[Dosovitskiy et~al.(2017)Dosovitskiy, Ros, Codevilla, Lopez, and Koltun]{Dosovitskiy17}
Alexey Dosovitskiy, German Ros, Felipe Codevilla, Antonio Lopez, and Vladlen Koltun.
\newblock {CARLA}: {An} open urban driving simulator.
\newblock In \emph{CoRL}, 2017.

\bibitem[Fitts(1967)]{fitts1967human}
PM~Fitts.
\newblock Human performance.
\newblock \emph{Brooks/Cole}, 1967.

\bibitem[Dallas et~al.(2023)Dallas, Thompson, Goh, and Balachandran]{DBLP:journals/fr/DallasTGB23}
James Dallas, Michael Thompson, Jonathan Y.~M. Goh, and Avinash Balachandran.
\newblock A hierarchical adaptive nonlinear model predictive control approach for maximizing tire force usage in autonomous vehicles.
\newblock \emph{Field Robotics}, 3\penalty0 (1):\penalty0 222--242, 2023.
\newblock \doi{10.55417/FR.2023006}.
\newblock URL \url{https://doi.org/10.55417/fr.2023006}.

\end{thebibliography}

\end{document}